\documentclass{article}

\usepackage{CJK}
\usepackage{graphicx}       % 核心：插入图片
\usepackage{caption}        % 可选：优化整体标题格式（推荐）
\usepackage{subcaption}     % 可选：用于多图的独立子标题（方案二需用）
\usepackage{amssymb}  % 提供 \checkmark 符号（√）
\usepackage{array}     % 优化表格格式（可选，用于行间距）
\usepackage{amsmath}

\usepackage{arxiv}

\usepackage[utf8]{inputenc} % allow utf-8 input
\usepackage[T1]{fontenc}    % use 8-bit T1 fonts
\usepackage{hyperref}       % hyperlinks
\usepackage{url}            % simple URL typesetting
\usepackage{booktabs}       % professional-quality tables
\usepackage{amsfonts}       % blackboard math symbols
\usepackage{nicefrac}       % compact symbols for 1/2, etc.
\usepackage{microtype}      % microtypography
\usepackage{lipsum}
\usepackage{graphicx}
\graphicspath{ {./images/} }

\title{CWSSNet: Hyperspectral Image Classification Enhanced by Wavelet Domain Convolution}

\author{
 Tong Yulin \\
  School of Transportation Engineering\\
  East China Jiaotong University\\
  Nanchang, Jiangxi  330013 \\
  \texttt{2019011008000108@ecjtu.edu.cn} \\
   \And
 Zhang Fengzong \\
  School of Transportation Engineering\\
  East China Jiaotong University\\
  Nanchang, Jiangxi  330013 \\
  \texttt{} \\
  \And
 Cheng Haiqin \\
  School of Coumputing and Information\\
  East China Jiaotong University\\
  Nanchang, Jiangxi  330013 \\
  \texttt{} \\
}

\begin{document}
\maketitle
\begin{abstract}
Hyperspectral remote sensing technology has significant application value in fields such as forestry ecology and precision agriculture, while also putting forward higher requirements for fine ground object classification. However, although hyperspectral images are rich in spectral information and can improve recognition accuracy, they tend to cause prominent feature redundancy due to their numerous bands, high dimensionality, and spectral mixing characteristics. To address this, this study used hyperspectral images from the ZY1F satellite as a data source and selected Yugan County, Shangrao City, Jiangxi Province as the research area to perform ground object classification research. A classification framework named CWSSNet was proposed, which integrates 3D spectral-spatial features and wavelet convolution. This framework integrates multimodal information us-ing a multiscale convolutional attention module and breaks through the classification performance bottleneck of traditional methods by introducing multi-band decomposition and convolution operations in the wavelet domain. The experiments showed that CWSSNet achieved 74.50\%, 82.73\%, and 84.94\% in mean Intersection over Union (mIoU), mean Accuracy (mAcc), and mean F1-score (mF1) respectively in Yugan County. It also obtained the highest Intersection over Union (IoU) in the classifica-tion of water bodies, vegetation, and bare land, demonstrating good robustness. Additionally, when the training set proportion was 70\%, the increase in training time was limited, and the classification effect was close to the optimal level, indicating that the model maintains reliable performance under small-sample training conditions.
\end{abstract}

% keywords can be removed

\section{Introduction}

Hyperspectral Image (HSI) refers to a type of remote sensing data acquired by imaging spectrometers through continuous imaging in the wavelength range from visible light to short-wave infrared, with a spectral resolution at the nanometer level. By combining the spatial characterization capability of imaging sensors and the spectral separation detection technology of spectrometers, HSI enables imaging of target areas within narrow spectral intervals. Consequently, each pixel is provided with dozens to hundreds of spectral bands, forming a continuous and complete spectral curve. This "image-spectrum integration" data structure significantly enhances the ability for fine-grained identification of ground object properties and quantitative inversion, providing crucial support for fields such as ground object classification \cite{1}, component analysis \cite{2}, and environmental monitoring \cite{3} \cite{4}.

Compared with traditional remote sensing images with low spectral resolution, HSI exhibits the following distinctive characteristics: (1) A substantial increase in the number of bands and narrow spectral band width, which enhances the separability of spectral features of ground objects \cite{5}; (2) The capability to capture more subtle differences in the reflection/radiation properties of ground objects, thereby revealing more abundant information about ground object attributes \cite{6}; (3) High spectral continuity, which facilitates the differentiation of subcategories within the same ground object type or identical ground objects in different states \cite{7}; (4) Enormous data volume——due to the increased number of bands and the expanded information dimension contained in each pixel, the overall data scale of HSI is significantly enlarged, imposing higher requirements on storage and computation \cite{8}.

In recent years, the processing and analysis of HSI have become a key research topic in the field of remote sensing. Technological breakthroughs in this area serve as a critical foundation for realizing the transition from qualitative discrimination to quantitative analysis, and also represent a core challenge to be addressed urgently in current remote sensing application research. Meanwhile, scholars at home and abroad have conducted extensive studies in the field of remote sensing image semantic segmentation, achieving remarkable progress. Traditional methods mostly rely on manually designed feature extractors and classifiers (e.g., Support Vector Machines, Random Forests). Although these methods can achieve ground object classification to a certain extent, they still face challenges such as limited feature expression capability and insufficient classification accuracy \cite{9}. With the development of deep learning technology, semantic segmentation methods based on Convolutional Neural Networks (CNN) have gradually become a research focus \cite{10}. Owing to their hierarchical structure, CNNs can effectively learn multi-level features, significantly improving classification performance \cite{11}. For instance, Chen et al. \cite{12} applied 3D-CNN for HSI classification; however, texture category confusion still exists in certain spectral bands, which limits the further improvement of classification accuracy. The HybridSN model proposed by Swalpa et al. \cite{13} integrates 2D-CNN and 3D-CNN layers, leveraging both spectral and spatial information to achieve favorable classification results. Nevertheless, such methods are constrained by local receptive fields, making it difficult to effectively model the semantic correlations among ground objects over a large range. If traditional convolution operations expand the receptive field by increasing the convolution kernel size, the number of parameters will grow quadratically. This not only increases the computational burden but also easily leads to overfitting issues \cite{14}.

To strike a balance between segmentation performance and computational efficiency \cite{15}, and to bridge the gap between local semantic expression and global scene understanding \cite{16}, this paper proposes a remote sensing image semantic segmentation network based on CNN and Multi-scale Wavelet Transform (WT), namely the CNN-Wavelet based Semantic Segmentation Network (CWSSNet). This network incorporates an attention mechanism to refine the feature outputs of the encoder \cite{17}, thereby fusing local details \cite{18} and global contextual information \cite{19}. The CWSSNet adopts an encoder-decoder structure: the encoder takes CNN as the backbone, embeds a Multi-scale Conv Attention (MCA) module to extract multi-dimensional features, and is supplemented with a wavelet convolution module to enhance multi-frequency and multi-scale feature representation; the decoder integrates multi-level information through a feature fusion module. The main contributions of this paper are as follows:

(1) A semantic segmentation network (CWSSNet) that integrates CNN and WT is proposed. It can effectively capture local-global contextual information with low parameter counts and computational costs, realize multi-scale feature fusion, and thus improve segmentation accuracy.

(2) A Multi-scale Conv Attention (MCA) module is designed. It employs 3D convolution to process HSI multi-dimensional data for exploring spatial-spectral correlations, and realizes the fusion and weighting of local and global features through a dual-pooling branch, thereby effectively suppressing redundant information and integrating multi-modal features.

(3) A Wavelet Threshold Binary Convolution (WTBC) module is proposed. Binary convolution operations and a multi-scale feature interaction mechanism are introduced into traditional wavelet convolution to efficiently aggregate global semantics and local details, enhancing the overall accuracy of semantic segmentation and the ability to restore details.

(4) A Feature Fusion module is constructed. With the help of a dual-branch attention mechanism, it screens effective features from two dimensions (global statistics and local extrema), thereby promoting the generation of more refined and accurate segmentation results.

\section{Research Methods}
\label{sec:headings}
The overall architecture of CWSSNet is illustrated in Figure 1. The input to the network is a hyperspectral remote sensing image with a size of M×N×D, where M×N represents the spatial dimensions and D denotes the original spectral dimension. First, Principal Component Analysis (PCA) dimensionality reduction is performed on the input data, reducing the number of spectral bands from D to B, resulting in data with a dimension of M×N×B. This step eliminates redundant spectral information and focuses on key features. Subsequently, the dimensionally reduced data is divided into overlapping 3D image patches P with a size of S×S×B by means of image classification techniques. This process reduces spatial complexity while providing a more compact feature representation for the encoder, thereby achieving a good balance between computational efficiency and the level of feature abstraction.

\begin{figure}[htbp]  % 建议添加[htbp]参数，优化图片排版位置
    \centering
    % 临时设置当前figure的标题前缀为"Fig."，编号格式为"Fig. 1"
    \captionsetup{name=Fig., labelsep=space}  % labelsep=space：编号与标题文字间用空格分隔
    \includegraphics{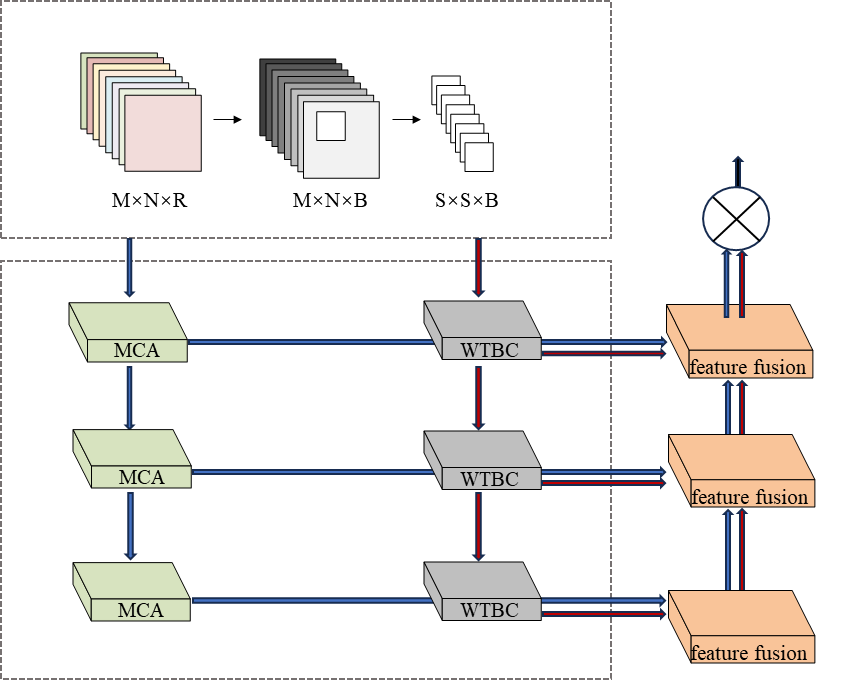}  % 你的图片文件
    \caption{Structure of CWSSNet}  % 标题内容（无需手动写编号，LaTeX自动生成）
    \label{fig:cwssnet_structure}  % 可选：添加标签，用于后文引用（如"Fig. \ref{fig:cwssnet_structure}"）
\end{figure}

The encoder adopts Convolutional Neural Network (CNN) as its basic framework, which serially connects the Multi-Channel Attention module (MCA) and the Wavelet Threshold Binary Convolution module (WTBC) in sequence. The MCA module utilizes 3D convolution to explore fine-grained features in the spatial-spectral dimensions, and employs a multi-scale pooling attention mechanism to highlight key information while suppressing redundant responses. The WTBC module, deployed in a parallel branch, first decomposes features into high-frequency details and low-frequency components through multi-scale Wavelet Transform (WT), then performs lightweight feature extraction via Binary Convolution (BC), and finally reconstructs multi-scale features using Inverse Wavelet Transform (IWT). This design effectively compensates for the deficiency of traditional CNNs in capturing multi-frequency features, forming a complement to the "global-local attention features" extracted by the MCA module.

The decoding part realizes the integration of cross-scale local details and global structures through a feature fusion module. The MCA features and WTBC features output by the encoder are input into the fusion module according to their scales. Each layer sequentially performs the following operations: 1×1 convolution for dimensionality reduction; multi-branch pooling (average pooling and max pooling) to extract multi-modal contextual features; MLP for further feature refinement with attention weighting via the Sigmoid function; and finally feature fusion through the Add operation. This strategy effectively reduces the hierarchical gap between the shallow details of the encoder and the deep semantics of the decoder, improving the spatial consistency and detail coherence of the segmentation results. Ultimately, the multi-layer fused features generate segmentation feature maps through point-wise multiplication, and the final semantic segmentation results are output through the classification head.

\paragraph{Task modeling.}
We approach this task as a regression problem. For every item and shop pair, we need to predict its next month sales(a number).
\paragraph{Construct train and test data.}
In the Sales train dataset, it only provides the sale within one day, but we need to predict the sale of next month. So we sum the day's sale into month's sale group by item, shop, date(within a month).
In the Sales train dataset, it only contains two columns(item id and shop id). Because we need to provide the sales of next month, we add a date column for it, which stand for the date information of next month.

\subsection{WTBC}
The core idea of the Wavelet Threshold Binary Convolution module (WTBC) is to combine the multi-scale analysis capability of Wavelet Transform (WT) with the efficient computation advantage of Binary Convolution (BC). While preserving spatial local details, it expands the receptive field with logarithmic complexity and significantly reduces computational overhead through binarization operations, thereby meeting the design requirements of lightweight models. The structures of BC and WTBC modules are shown in Figures 2 and 3, respectively. First, wavelet decomposition is performed on the input features to separate low-frequency and high-frequency components, which are then subjected to convolution processing at different scales: a 5×5 convolution is applied to the low-frequency components, while a 3×3 convolution is used for the high-frequency components. Subsequently, frequency attention weights are generated through the Sigmoid function to dynamically weight the convolution results of high and low frequencies, and feature fusion is achieved through point-wise multiplication.

\begin{figure}[htbp]  % 建议添加[htbp]参数，优化图片排版位置
    \centering
    % 临时设置当前figure的标题前缀为"Fig."，编号格式为"Fig. 1"
    \captionsetup{name=Fig., labelsep=space}  % labelsep=space：编号与标题文字间用空格分隔
    \includegraphics{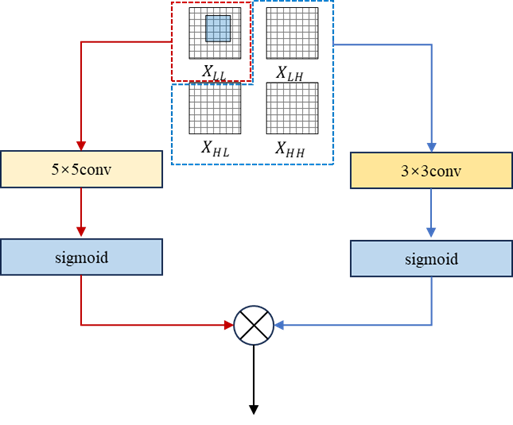}  % 你的图片文件
    \caption{Structure of Binary Convolution}  % 标题内容（无需手动写编号，LaTeX自动生成）
    \label{fig:cwssnet_structure}  % 可选：添加标签，用于后文引用（如"Fig. \ref{fig:cwssnet_structure}"）
\end{figure}

\begin{figure}[htbp]  % 建议添加[htbp]参数，优化图片排版位置
    \centering
    % 临时设置当前figure的标题前缀为"Fig."，编号格式为"Fig. 1"
    \captionsetup{name=Fig., labelsep=space}  % labelsep=space：编号与标题文字间用空格分隔
    \includegraphics{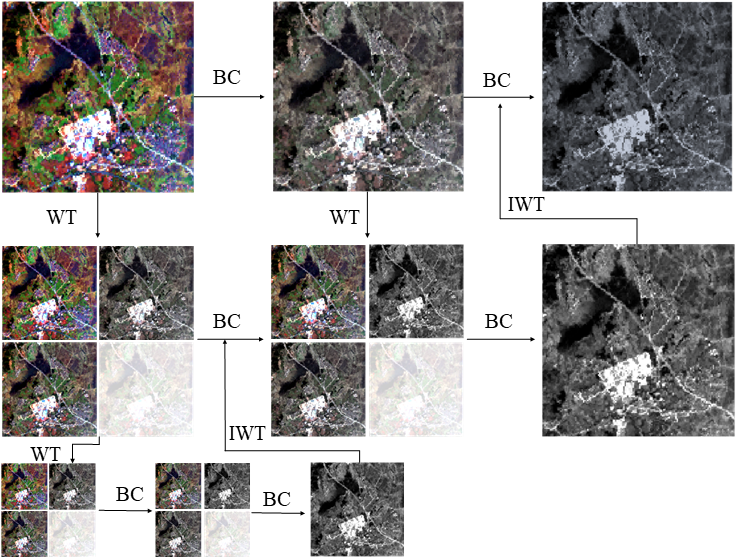}  % 你的图片文件
    \caption{Structure of WTBC Network}  % 标题内容（无需手动写编号，LaTeX自动生成）
    \label{fig:cwssnet_structure}  % 可选：添加标签，用于后文引用（如"Fig. \ref{fig:cwssnet_structure}"）
\end{figure}

Specifically, the wavelet decomposition at the i-th layer can be expressed as follows:

\begin{equation}
X_{\text{LL}}^{(i)},\ X_{\text{H}}^{(i)}\ =\ \text{WT}\left(X_{\text{LL}}^{(i-1)}\right)
\label{eq:wavelet_transform}  % 可选：添加标签，用于后文引用（如“见公式\ref{eq:wavelet_transform}”）
\end{equation}

Among them, \( X_{\text{LL}}^{(0)} \) denotes the initial input, and \( X_{\text{H}}^{(i)} \) represents all high-frequency sub-bands decomposed in the i-th layer. Subsequently, binary convolution is performed on the low-frequency and high-frequency components respectively:

\begin{equation}
Y_{\text{LL}}^{(i)}, Y_{\text{H}}^{(i)} = \text{BC}\left(W^{(i)}, \left(X_{\text{LL}}^{(i)}, X_{\text{H}}^{(i)}\right)\right)
\label{eq:bc_operation}
\end{equation}

Among them, \( W^{(i)} \) is the convolutional weight tensor (with a size of 3×3 or 5×5), and the number of output channels is four times that of the input. This operation not only achieves convolutional separation between frequency domains but also enables smaller convolutional kernels to function within a larger input region, thereby expanding their receptive fields \cite{20}. Further, the attention mechanism is used to weight different frequency components:

\begin{equation}
\text{Attn}_{\text{LL}} = \text{Sigmoid}\left(\text{Conv}_{5 \times 5}\left(X_{\text{H}}\right)\right)
\label{eq:attn_ll}
\end{equation}

\begin{equation}
\text{Fused}_{\text{LL}}, \text{Fused}_{\text{H}} = \text{Attn}_{\text{LL}} \otimes Y_{\text{LL}}^{(i)}, \text{Attn}_{\text{H}} \otimes Y_{\text{H}}^{(i)}
\label{eq:fused_calc}
\end{equation}

By leveraging the property that both wavelet transform (WT) and its inverse are linear operations (i.e., \( \text{IWT}(X + Y) = \text{IWT}(X) + \text{IWT}(Y) \)), the weighted multi-frequency features are integrated via inverse wavelet transform (IWT) to restore the spatial structure and fuse multi-scale information. This process aggregates the outputs of different layers in a recursive manner:

\begin{equation}
Z^{(i)} = \text{IWT}\left(\text{Fused}_{\text{LL}}^{(i)} + Z^{(i+1)}, \text{Fused}_{\text{H}}^{(i)}\right)
\label{eq:iwt_operation}
\end{equation}

Among them, \( Z^{(i)} \) denotes the aggregated output at the start of the i-th layer.

\subsection{MCA}
The structure of the Multi-Channel Attention module (MCA) is illustrated in Figure 4. First, the input features pass through a layer of 3D convolution to simultaneously explore the associated information in the spatial-spectral dimensions. Subsequently, batch normalization (BN) is applied to accelerate the training convergence, and a ReLU activation function is used to introduce non-linearity—thereby providing basic features rich in multi-dimensional semantics for the subsequent attention mechanism \cite{21}.

\begin{figure}[htbp]  % 建议添加[htbp]参数，优化图片排版位置
    \centering
    % 临时设置当前figure的标题前缀为"Fig."，编号格式为"Fig. 1"
    \captionsetup{name=Fig., labelsep=space}  % labelsep=space：编号与标题文字间用空格分隔
    \includegraphics{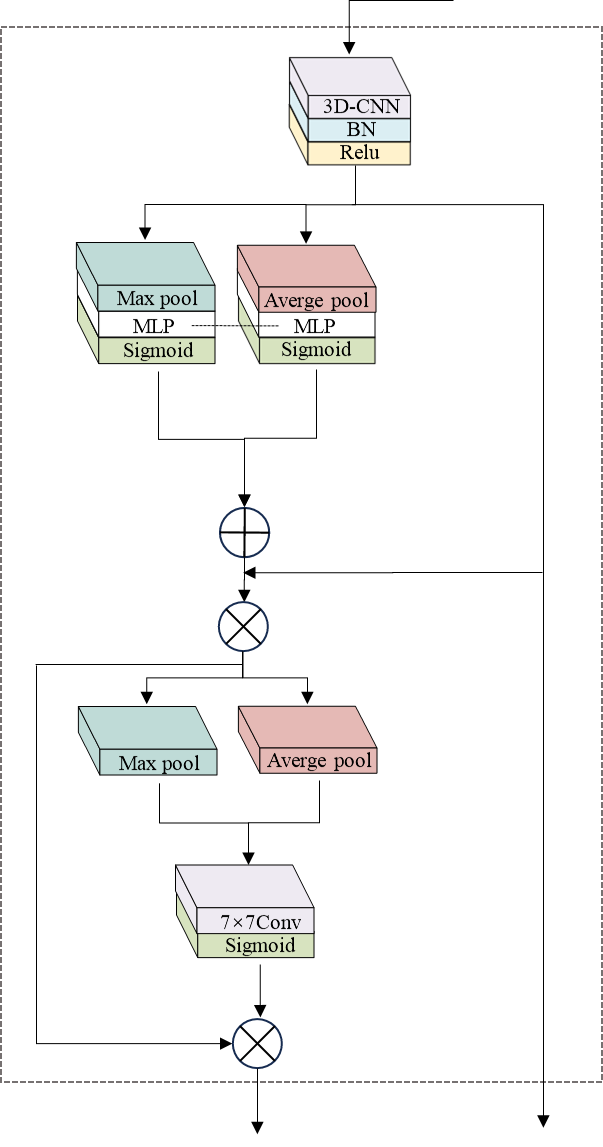}  % 你的图片文件
    \caption{Structure of MCA Network}  % 标题内容（无需手动写编号，LaTeX自动生成）
    \label{fig:cwssnet_structure}  % 可选：添加标签，用于后文引用（如"Fig. \ref{fig:cwssnet_structure}"）
\end{figure}

This module adopts a dual-branch parallel structure. One branch uses global average pooling to extract global statistical information of features, aiming to capture feature patterns within a wide range; the other branch focuses on local significant responses through max pooling to highlight extreme value features. These two are respectively input into a Multi-Layer Perceptron (MLP), which compresses dimensions while learning feature interaction relationships, and generates attention weights through the Sigmoid function respectively. The weights of the two branches are summed to fuse global statistics and local prominent features, and then point-wise multiplied with the output features of the 3D convolution to achieve preliminary feature reweighting, enabling the model to focus on key content filtered through multi-scale pooling. This process can be expressed as:

\begin{equation}
A = \text{MLP}\left(\text{AvgPool}\left(F_{3d}\right)\right)
\label{eq:mlp_avgpool}
\end{equation}

\begin{equation}
B = \text{MLP}\left(\text{MaxPool}\left(F_{3d}\right)\right)
\label{eq:mlp_maxpool}
\end{equation}

\begin{equation}
F_{\text{mid1}} = \left[\sigma\left(A\right) + \sigma\left(B\right)\right] \otimes F_{3d}
\label{eq:f_mid1_calc}
\end{equation}

Here, \( F_{3d} \) denotes the features processed by ReLU, \( \sigma \) represents the Sigmoid function, and \( \otimes \) stands for point-wise multiplication.  

The weighted feature \( F_{\text{mid1}} \) undergoes global average pooling and max pooling again. The results of these two operations are summed and then passed through a 7×7 convolutional layer, followed by the Sigmoid function to generate global attention weights. This large-scale convolution kernel can cover a broader spatial-spectral region, effectively modeling long-range dependencies. Finally, the generated weights are point-wise multiplied with \( F_{\text{mid1}} \) to further suppress redundant features and enhance highly discriminative features, forming a multi-scale feature screening and enhancement mechanism from local to global \cite{22}. This step can be expressed as:

\begin{equation}
C = \text{Conv}_{7 \times 7}\left(\text{AvgPool}\left(F_{\text{mid1}}\right) + \text{MaxPool}\left(F_{\text{mid1}}\right)\right)
\label{eq:conv_7x7}
\end{equation}

\begin{equation}
F_{\text{out}} = \sigma\left(C\right) \otimes F_{\text{mid1}}
\label{eq:f_out_calc}
\end{equation}

The output feature \( F_{\text{out}} \) undergoes attention weighting intervention twice, completing multi-level feature refinement and providing a more discriminative representation for downstream tasks.

\subsection{Feature Fusion}
Due to the differences in design logic between the WTBC and MCA modules, the generated features also exhibit representations across distinct dimensions: The WTBC module relies on wavelet multi-scale decomposition and binary convolution, focusing on extracting high-frequency detail features from multi-frequency sub-bands. In contrast, the MCA module leverages 3D-CNN to explore spatial-spectral dimensional correlations, and strengthens key semantic information through multi-scale pooling and an attention mechanism while suppressing redundant responses. If the features output by these two modules are directly concatenated, the differences in feature levels and representation modes will lead to semantic aliasing, making it difficult to achieve full fusion of effective information.

\begin{figure}[htbp]  % 建议添加[htbp]参数，优化图片排版位置
    \centering
    % 临时设置当前figure的标题前缀为"Fig."，编号格式为"Fig. 1"
    \captionsetup{name=Fig., labelsep=space}  % labelsep=space：编号与标题文字间用空格分隔
    \includegraphics{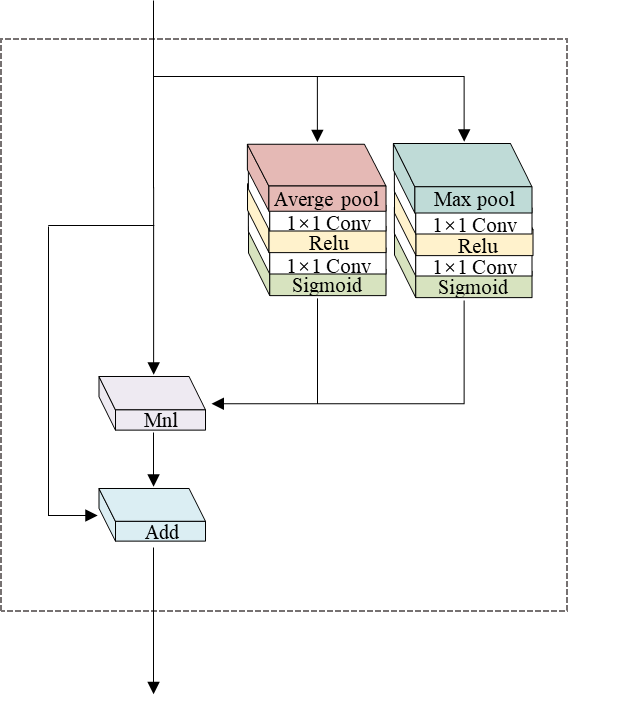}  % 你的图片文件
    \caption{Feature Fusion Module}  % 标题内容（无需手动写编号，LaTeX自动生成）
    \label{fig:cwssnet_structure}  % 可选：添加标签，用于后文引用（如"Fig. \ref{fig:cwssnet_structure}"）
\end{figure}

To address this issue, a feature fusion module is designed in this study (its workflow is shown in Figure 5). First, this module performs max pooling and average pooling operations on the input features respectively. Subsequently, two 1×1 convolutional layers are employed to control the model complexity, with a ReLU activation function embedded between the convolutions to enhance the nonlinear expression capability. Finally, an attention weight vector is generated through the Sigmoid function. The output of the two attention weight branches is interacted with the features further refined by the MLP, and the features processed by different branches (including attention weighting information) are fused via the Add operation. Eventually, the integrated enhanced features are output.

\section{Experiment}
\subsection{Overview of the Experimental Area}
The study area is located in Yugan County, Shangrao City, which is situated in the northeastern part of Jiangxi Province, on the southeastern bank of Poyang Lake, and in the lower reaches of the Xin River. Its geographical coordinates range from 116°13'48"E to 116°54'24"E in longitude and from 28°21'36"N to 29°3'24"N in latitude \cite{23}. Yugan County is adjacent to Wannian County in the east, Yujiang District of Yingtan City and Dongxiang District of Fuzhou City in the south, Jinxian County, Xinjian District, and Nanchang County of Nanchang City in the west, and Poyang County in the north; it also faces Duchang County across Poyang Lake. The county presents a narrow and long shape from north to south, with a north-south length of approximately 87 km, an east-west width of about 38 km, and a total area of 2336 km².
The main landforms in the area are low hills and lakeside plains, with the terrain sloping from southeast to northwest, transitioning gradually from hills to plains. The highest point is Limei Ridge, with an altitude of 390 meters, while the lowest point is Huangdihmao in the north, with an altitude of approximately 13 meters.

The remote sensing data used in this study were acquired from the ZY1F satellite on January 24, 2023. The data have a spatial resolution of 30 meters and include a total of 150 spectral bands. The original image product is at Level L1, which has undergone preprocessing such as bad pixel repair, radiometric correction, and spectral correction. However, further orthorectification and atmospheric correction are still required \cite{24}.

To this end, the following preprocessing steps were implemented:

Radiometric calibration: The original Digital Number (DN) values of the images were converted into radiance values.
Atmospheric correction: The FLAASH module in ENVI software was used to perform atmospheric correction, so as to obtain the true reflectance data of ground objects.

The preprocessed image of the experimental area is shown in the figure (with false-color synthesis using bands 56, 36, and 26).

\subsection{Experimental Setup}
The experiments were conducted on the Windows 10 operating system and implemented based on the Python 3.10.6 environment and PyTorch 2.2.1 framework. The AdamW optimizer was adopted, with a base learning rate set to 0.006.

To evaluate the segmentation performance of the proposed network, three evaluation metrics were used, including mean Intersection over Union (mIoU), mean F1 Score (mF1), and mean Accuracy (mAcc). Among them, mIoU, as a mainstream metric, measures segmentation accuracy \cite{25}; mF1 focuses on the model’s performance under class imbalance; and mAcc is used to evaluate the overall classification accuracy. The calculation formulas for each metric are as follows:

\begin{equation}
\text{mIoU} = \frac{1}{N} \sum_{k=1}^{N} \frac{TP_k}{TP_k + FP_k + FN_k}
\label{eq:miou}
\end{equation}

\begin{equation}
\text{mF1} = \frac{1}{N} \sum_{k=1}^{N} \frac{2 \cdot TP_k}{2 \cdot TP_k + FN_k + FP_k}
\label{eq:mf1}
\end{equation}

\begin{equation}
\text{mAcc} = \frac{1}{N} \sum_{k=1}^{N} \frac{TP_k}{TP_k + FN_k}
\label{eq:macc}
\end{equation}

Among them, \( N \) represents the total number of classes. \( \text{TP}_k \) (True Positives) denotes the number of pixels in the \( k \)-th class that are correctly predicted as belonging to this class. \( \text{FP}_k \) (False Positives) stands for the number of pixels that are incorrectly predicted as belonging to the \( k \)-th class but do not actually belong to it. \( \text{FN}_k \) (False Negatives) refers to the number of pixels that actually belong to the \( k \)-th class but are not correctly predicted as such.  

Pseudocode of the CWSSNet Hyperspectral Image Semantic Segmentation Algorithm  

Algorithm 1: CWSSNet Hyperspectral Image Semantic Segmentation Algorithm  

Input: Hyperspectral image patch (spatial size: \( S \times S \), number of bands: \( D \))  

Output: Semantic segmentation map (number of classes: \( C \))

Initialization Phase:

1. Data Preprocessing: Perform PCA dimensionality reduction on the input image patch to compress the original high-dimensional spectral data into main bands (resulting in \( B \) bands, \( B < D \)).  

2. Tensor Reshaping: Adjust the data dimension to \( 1 \times S \times S \times B \times 1 \) (add a pseudo-depth dimension to adapt to 3D convolution operations).  

3. Model Parameter Initialization: Initialize the weights of all convolutional layers using He initialization; set the parameters of batch normalization layers to the identity transformation.

Training Process: 

for each training epoch do:

Sample image patches from the training set:
    
Read batch data $P_{batch}$(batch size: N)
    
Apply PCA dimensionality reduction and adjust tensor dimension to obtain input tensor X 
    
// ===== First-Level Feature Extraction =====
    
Extract spatial-spectral joint features:
    
Use the multi-channel attention (MCA) mechanism to capture global context and generate refined features M1

Apply wavelet threshold binary convolution (WTBC) to decompose frequency-domain features and extract multi-scale detail features W1

Fuse dual-path features:
    
Combine the semantic information of M1 and texture details of W1 via the attention-guided feature fusion module to obtain fused feature F1
    
Spatial downsampling:
    
Apply max-pooling to F1 to generate low-resolution feature D1 (spatial size is halved, i.e., \( (S/2) \times (S/2) \))
    
// ===== Second-Level Feature Extraction =====
    
Deepen feature representation:
    
Re-apply the MCA mechanism on D1 to enhance deep semantic features M2

Perform secondary wavelet decomposition on D1 to extract more abstract frequency-domain representations W2
    
Fuse deep features:
    
Generate high-semantic feature F2 by fusing M2 and W2 through the feature fusion module
    
Secondary spatial compression:
    
Apply max-pooling to F2 to obtain highly compressed feature D2 (spatial size is reduced to 1/4 of the original, i.e., \( (S/4) \times (S/4) \))
    
// ===== Decoding and Reconstruction =====
    
Feature upsampling:
    
Perform deconvolution on D2 to restore spatial resolution, resulting in U1 (spatial size is doubled, i.e., \( (S/2) \times (S/2) \))
    
Skip connection fusion:
    
Concatenate U1 with the second-level fused feature F2, then apply convolution to fuse shallow and deep features to generate enhanced feature FU1

Secondary upsampling:
    
Perform deconvolution on FU1 to further restore resolution, resulting in U2 (spatial size is restored to the original \( S \times S \))
    
Final feature fusion:
    
Concatenate U2 with the first-level fused feature F1 to supplement shallow details and obtain reconstructed feature FU2
    
// ===== Prediction and Optimization =====

Pixel-level classification:
    
Pass FU2 through a 1×1 convolutional layer to generate classification logits

Apply the softmax activation function to obtain the pixel-wise class probability distribution $Y_{pred}$
    
Loss calculation:
    
Compute cross-entropy loss: \( \text{CE}(Y_{\text{pred}}, Y_{\text{true}}) \)

Add L2 regularization term: \( \lambda \| W \|^2 \) (where \( W \) denotes all trainable weights of the model, \( \lambda \) is the regularization coefficient)

Total loss = Cross-entropy loss + L2 regularization term
    
Backpropagation:
    
Compute gradients of the total loss with respect to model parametersUpdate weights using the Adam optimizer

end for

\subsection{Experimental Results and Analysis}
In this study, 4 traditional land cover classification methods and 3 deep learning-based land cover classification methods were tested on the study area. Table 1 presents the experimental results of each method, including the Intersection over Union (IoU) for each class and the overall evaluation metrics (mean IoU (mIoU), mean Accuracy (mAcc), and mean F1 Score (mF1)). The optimal results are highlighted in bold.
As shown in Table 1, among all comparative methods, the proposed CWSSNet achieves the optimal performance in the three evaluation metrics (mIoU, mAcc, mF1), reaching 74.50\%, 82.73\%, and 84.94\% respectively. Additionally, it obtains the best IoU results for classes such as water bodies, vegetation, and bare soil. Its classification results are visualized in Figure 6.
Among the classes (construction land, water bodies, vegetation, bare soil, roads, and others), vegetation and bare soil exhibit wide continuous distribution, irregular shapes, and similar spectral features in some regions. Distinguishing their subtle differences requires simultaneously capturing long-range dependencies and local fine-grained information. Taking vegetation and bare soil as examples, the limitations of comparative methods and the advantages of CWSSNet are analyzed as follows:

Limitations of Traditional Methods: Traditional methods such as Decision Tree and Support Vector Machine (SVM) rely on single-feature segmentation. When facing regions with similar spectral characteristics, they tend to cause confusion between vegetation and bare soil. Although Random Forest (RF) improves generalization ability by integrating multiple decision trees, it still lacks sufficient multi-scale feature extraction capabilities, making it difficult to capture the texture details of vegetation and the structural contours of bare soil.
Limitations of Deep Learning Methods: 2D CNN and 3D CNN are constrained by the local receptive field of convolution operations, resulting in weak modeling of long-range spatial-spectral dependencies. This leads to incomplete segmentation of large continuous vegetation or bare soil areas. HybridSN, which combines 2D and 3D convolution, achieves certain improvements in feature fusion but still has shortcomings in distinguishing details in complex regions with similar features—for instance, it may misclassify sparse vegetation edges as bare soil.
Advantages of CWSSNet:
WTBC Module for Detail Separation: The multi-scale wavelet decomposition in the Wavelet Threshold Binary Convolution (WTBC) module clearly separates the texture details of vegetation (e.g., leaf density variations) from the structural contours of bare soil (e.g., surface undulations). Through binary convolution and inverse wavelet transform fusion, it retains fine-grained information while reducing computational overhead—addressing the efficiency-accuracy trade-off.
MCA Module for Semantic Enhancement: The Multi-Channel Attention (MCA) module strengthens the semantic differences between vegetation and bare soil via multi-scale pooling attention. It assigns higher weights to discriminative features (e.g., vegetation’s unique spectral peaks in the near-infrared band, bare soil’s high reflectance in the visible band) and suppresses redundant noise, enhancing the model’s ability to distinguish similar spectral regions.
Feature Fusion Module for Cross-Level Aggregation: The feature fusion module narrows the semantic gap between the encoder (which extracts deep semantic features) and the decoder (which recovers spatial details). It effectively aggregates global semantic information (e.g., the overall distribution of vegetation communities) and local detail features (e.g., small bare soil patches within vegetation areas), ensuring both the accuracy of large-region segmentation and the coherence of edge details.

Thanks to these advantages, CWSSNet achieves IoU values of 79.45\% for vegetation and 85.63\% for bare soil, outperforming all comparative methods. This demonstrates its strength in fine-grained distinction between similar classes, effectively solving the classification challenges caused by the irregular distribution and similar spectral features of vegetation and bare soil, and improving the accuracy and detail coherence of semantic segmentation in complex scenarios.

\begin{figure}[htbp]
    \centering
    % 第1行
    \begin{subfigure}[b]{0.48\textwidth}
        \centering
        \includegraphics[width=\textwidth]{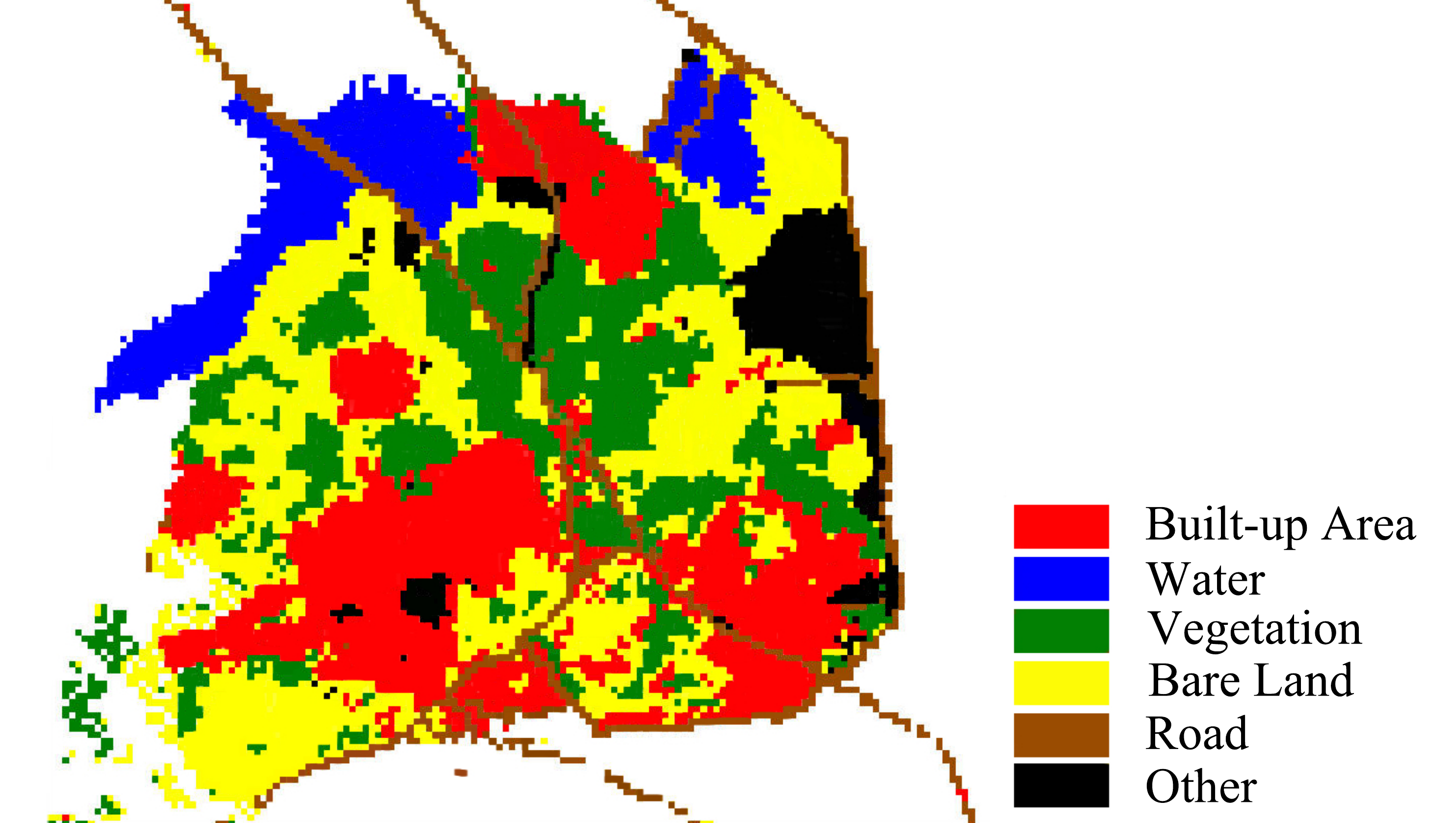}  % 第1张图
        \caption{Reference Diagram of Classification Results}
        \label{subfig:1}
    \end{subfigure}
    \hfill
    \begin{subfigure}[b]{0.48\textwidth}
        \centering
        \includegraphics[width=\textwidth]{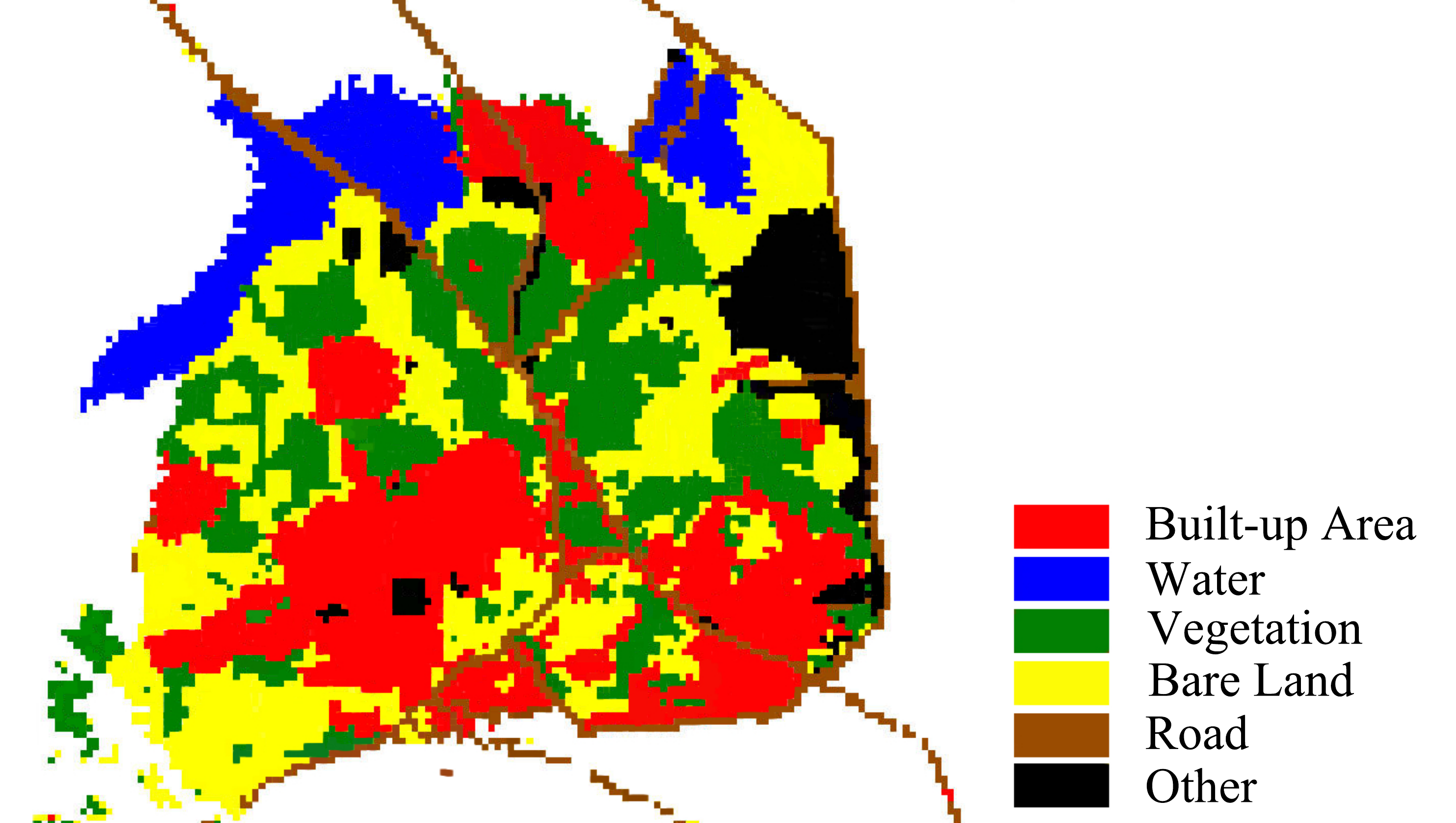}  % 第2张图
        \caption{CWSSNet}
        \label{subfig:2}
    \end{subfigure}
    
    % 第2行（用空行分隔不同行）
    \begin{subfigure}[b]{0.48\textwidth}
        \centering
        \includegraphics[width=\textwidth]{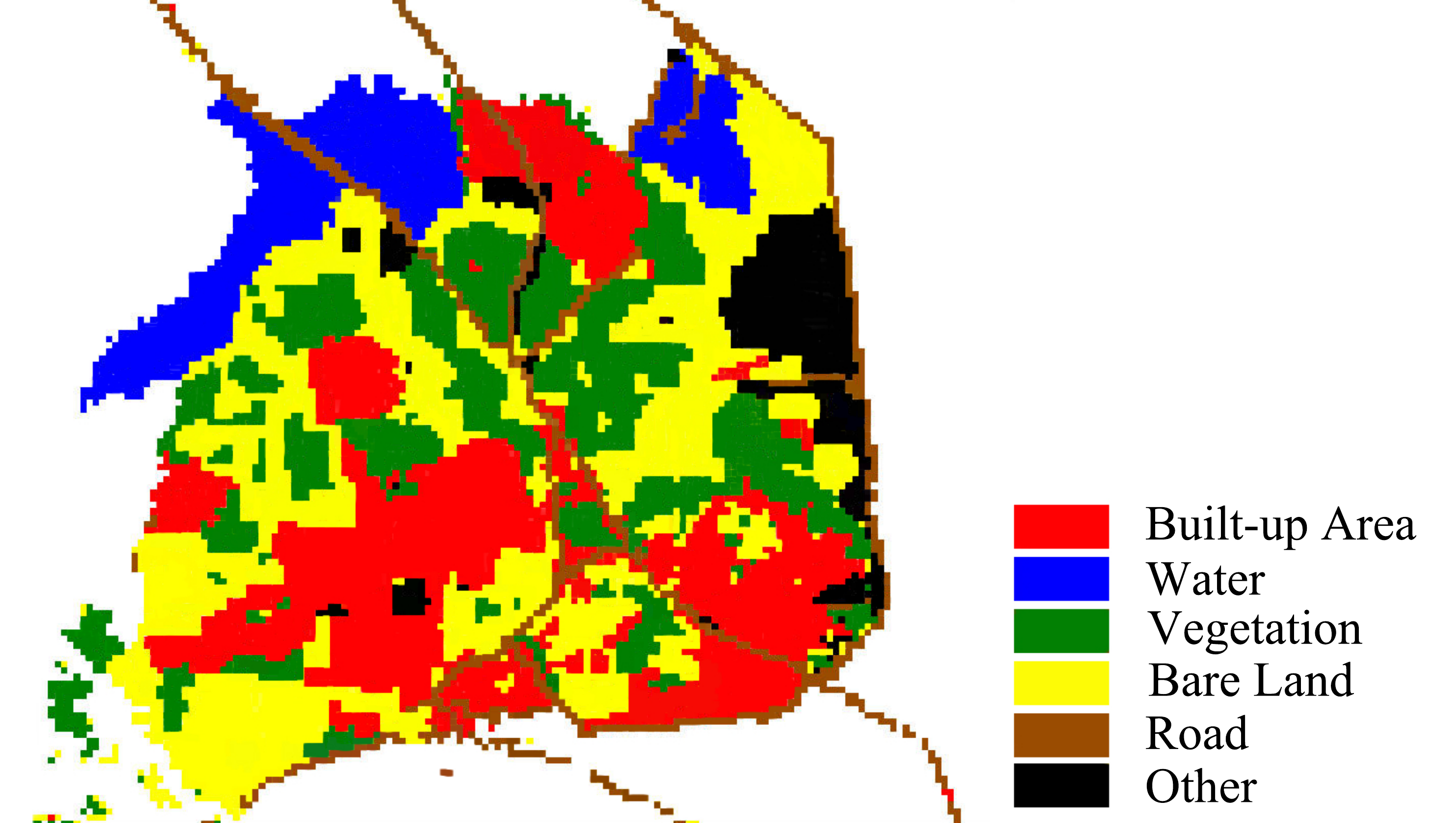}  % 第3张图
        \caption{RF}
        \label{subfig:3}
    \end{subfigure}
    \hfill
    \begin{subfigure}[b]{0.48\textwidth}
        \centering
        \includegraphics[width=\textwidth]{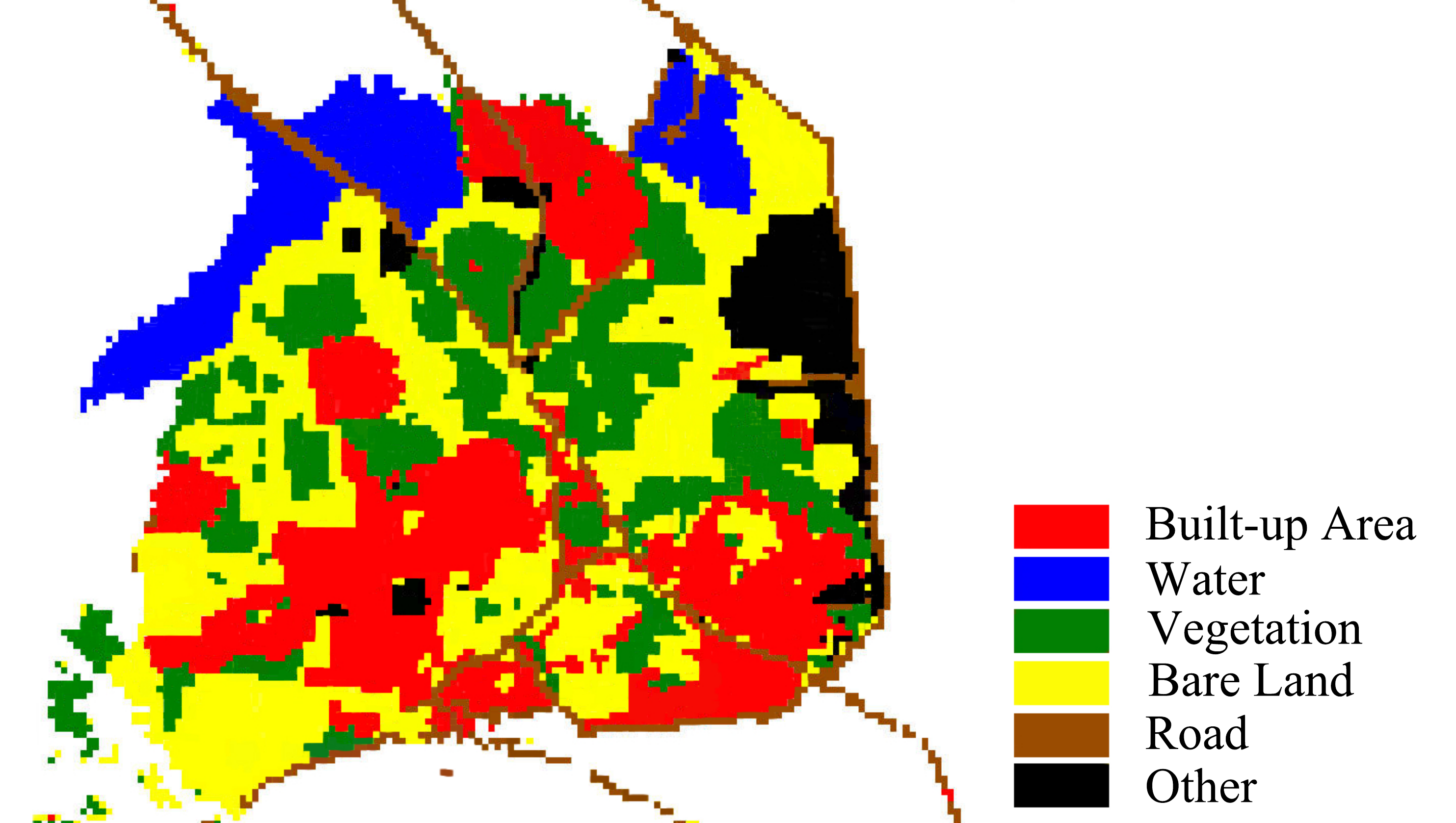}  % 第4张图
        \caption{GBDT}
        \label{subfig:4}
    \end{subfigure}
    
    % 第3行
    \begin{subfigure}[b]{0.48\textwidth}
        \centering
        \includegraphics[width=\textwidth]{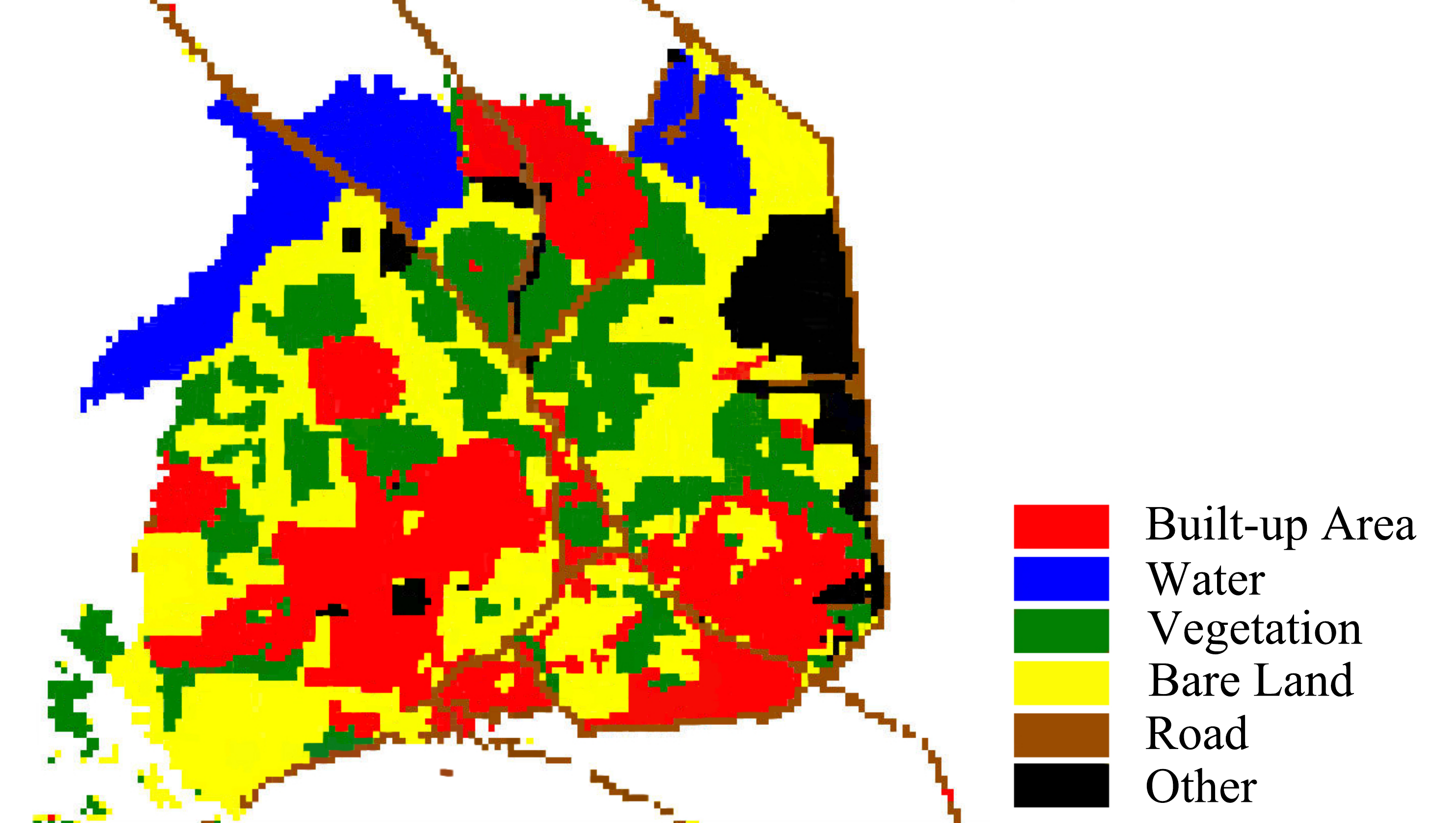}  % 第5张图
        \caption{SVM}
        \label{subfig:5}
    \end{subfigure}
    \hfill
    \begin{subfigure}[b]{0.48\textwidth}
        \centering
        \includegraphics[width=\textwidth]{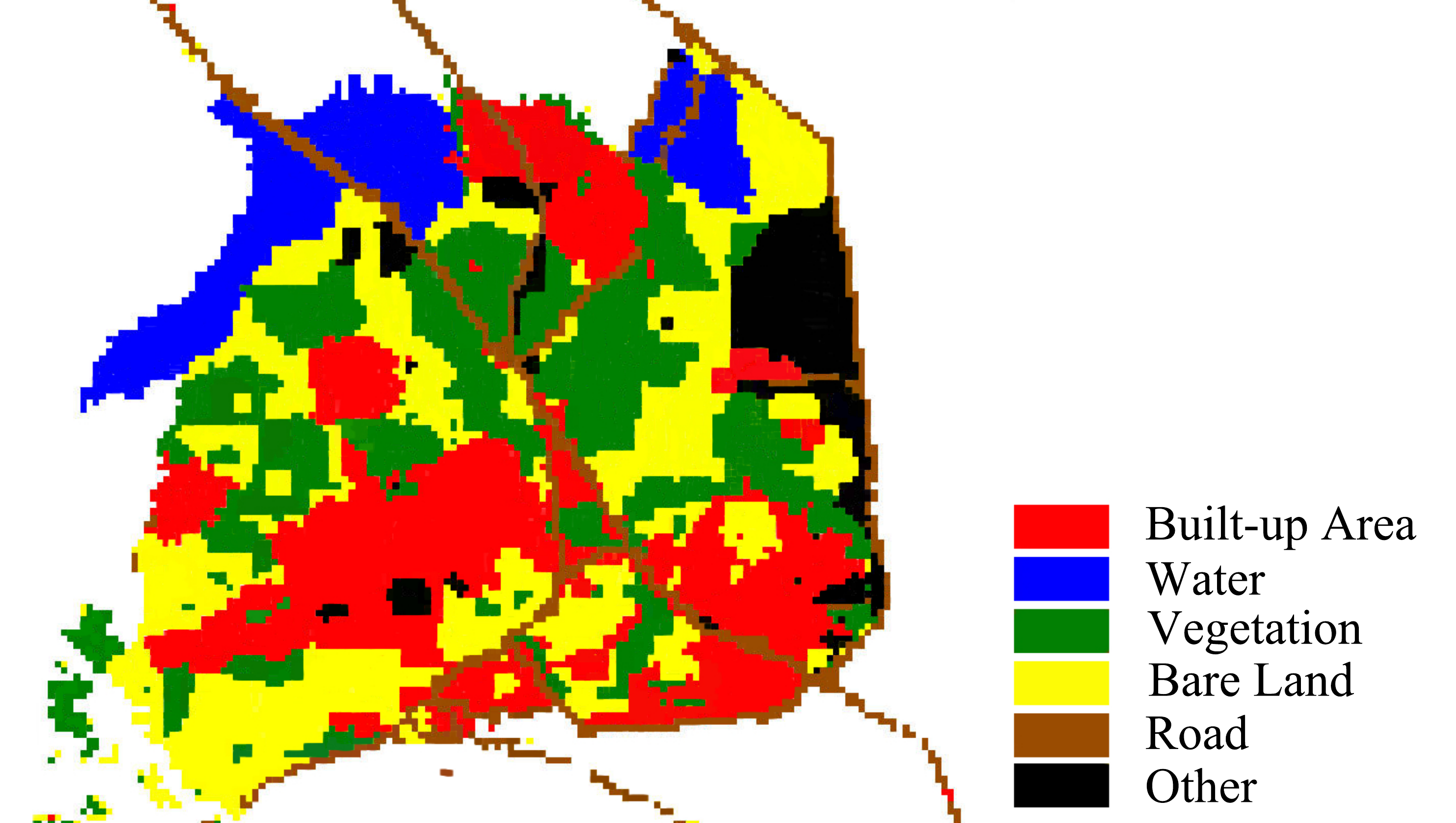}  % 第6张图
        \caption{2DCNN}
        \label{subfig:6}
    \end{subfigure}
    
    % 第4行
    \begin{subfigure}[b]{0.48\textwidth}
        \centering
        \includegraphics[width=\textwidth]{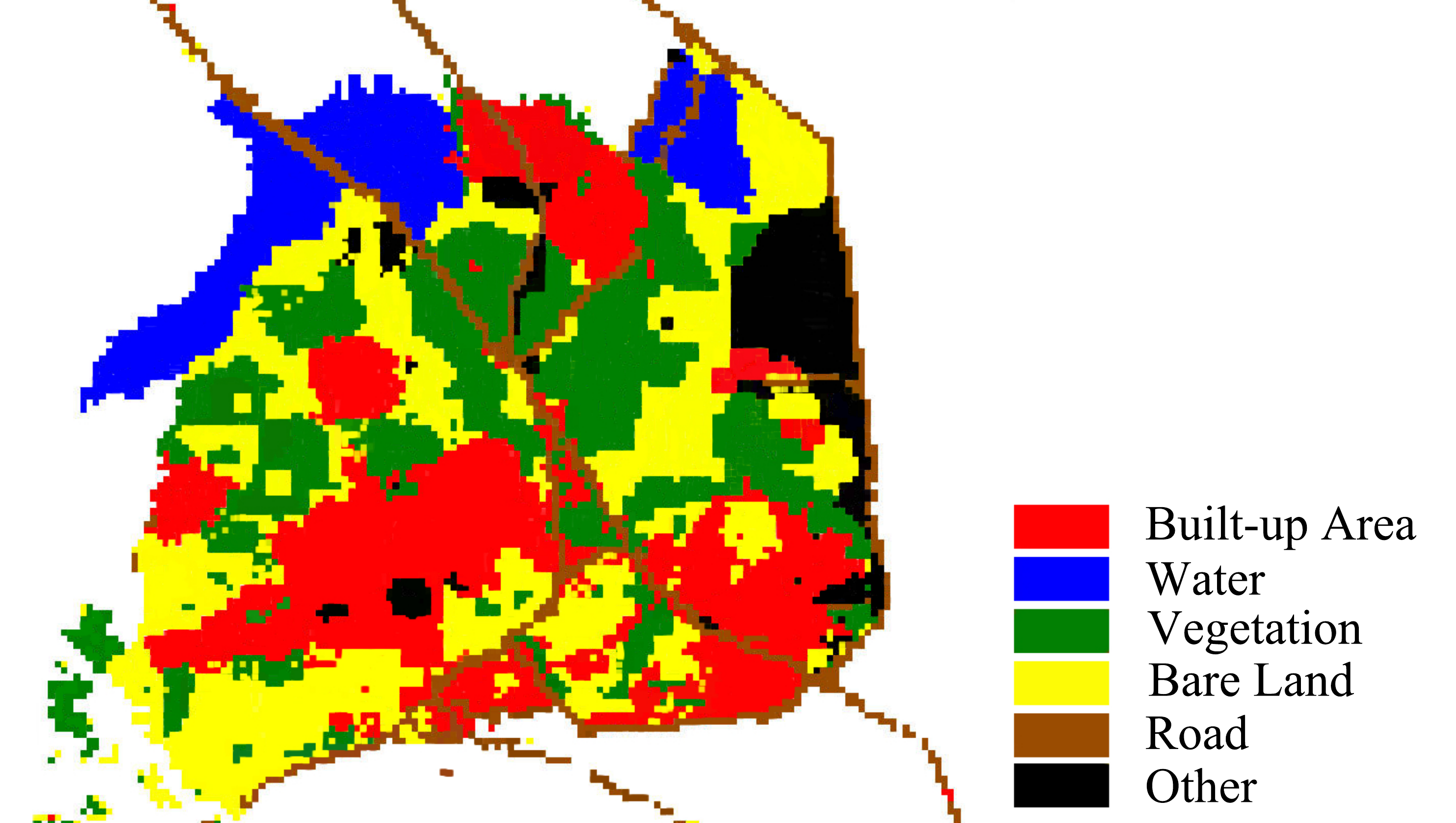}  % 第7张图
        \caption{3DCNN}
        \label{subfig:7}
    \end{subfigure}
    \hfill
    \begin{subfigure}[b]{0.48\textwidth}
        \centering
        \includegraphics[width=\textwidth]{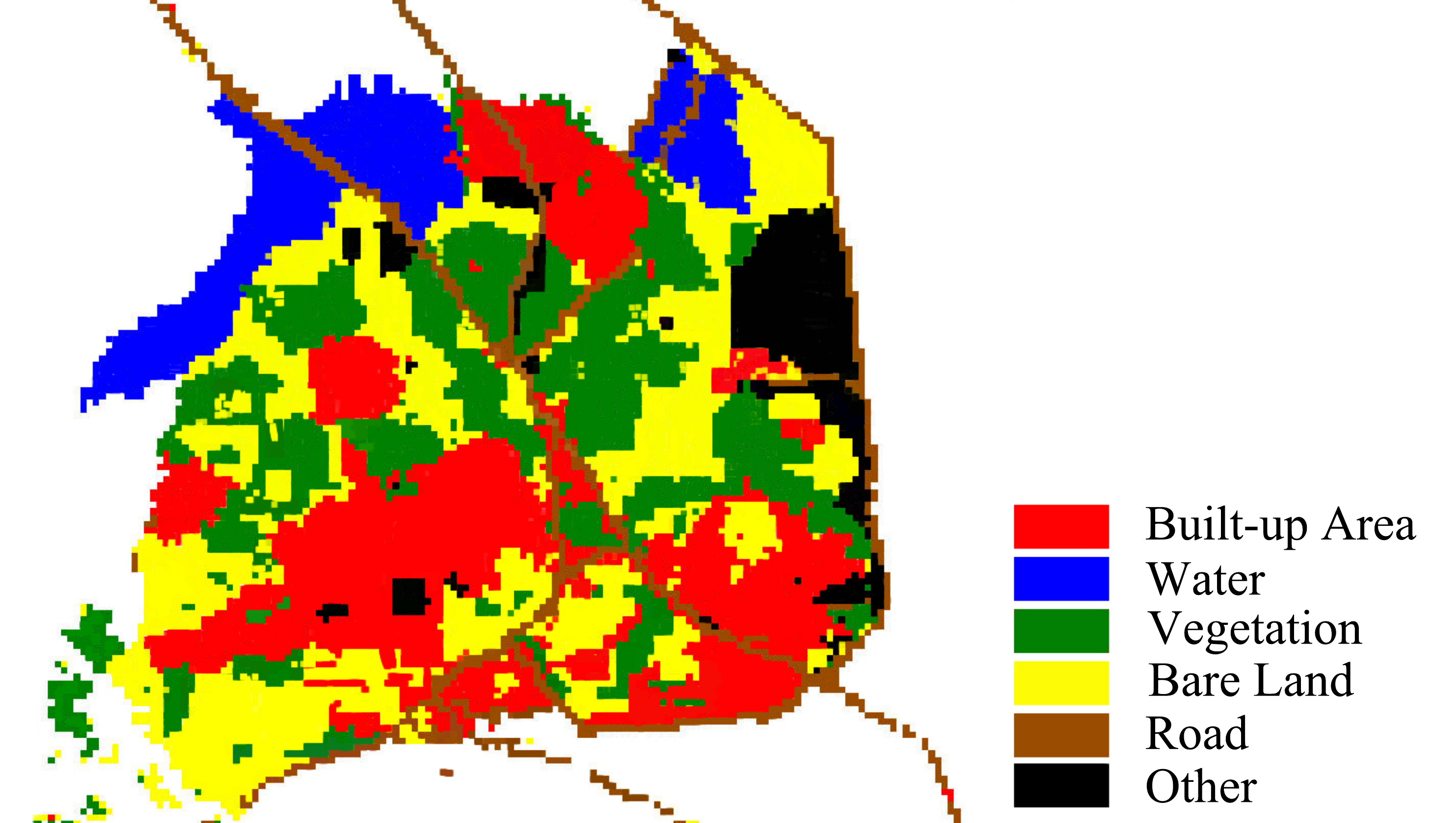}  % 第8张图
        \caption{HybridSN}
        \label{subfig:8}
    \end{subfigure}
    
    % 整体标题（所有子图共用）
    \caption{Classification Results of Various Algorithms}
    \label{fig:8images}
\end{figure}

\begin{table}[]
 \caption{Comparison of Classification Results of Various Algorithms}
\centering
  \setlength{\extrarowheight}{4pt}  % 增加行间距，避免内容拥挤（可选）
  % 列格式：8列居中对齐，每列两侧加竖线（|c| 表示带竖线的居中列）
\begin{tabular}{c|ccccccc}
\hline
\begin{tabular}[c]{@{}c@{}}Class\\ No.\end{tabular} & GBDT  & SVM   & RF    & 2DCNN & 3DCNN & HybridSN & CWSSNet \\ \hline
1                                                   & 85.48 & 85.64 & 85.65 & 85.61 & 86.80 & 85.81    & 86.09   \\
2                                                   & 88.18 & 87.81 & 86.56 & 91.58 & 91.63 & 91.20    & 91.87   \\
3                                                   & 71.26 & 71.33 & 72.29 & 71.83 & 70.04 & 75.43    & 79.45   \\
4                                                   & 74.95 & 73.86 & 74.73 & 77.02 & 81.38 & 83.63    & 85.63   \\
5                                                   & 39.19 & 39.46 & 39.10 & 38.73 & 40.30 & 44.19    & 45.21   \\
6                                                   & 65.16 & 64.51 & 66.40 & 68.78 & 66.75 & 70.21    & 71.25   \\ \hline
mloU                                                & 73.16 & 73.26 & 73.17 & 73.08 & 73.48 & 73.58    & 74.50   \\
mAcc                                                & 79.72 & 80.29 & 80.10 & 80.07 & 81.73 & 82.22    & 82.73   \\
mF1                                                 & 80.73 & 81.10 & 80.76 & 80.72 & 84.28 & 84.50    & 84.94   \\ \hline
\end{tabular}
\end{table}

The CWSSNet proposed in this study effectively alleviates the aforementioned issues through multi-module collaboration:

The WTBC (Wavelet Threshold Binary Convolution) module leverages multi-scale wavelet decomposition to effectively separate the texture details of vegetation from the structural contours of bare soil. Combined with binary convolution and inverse wavelet transform, it achieves both detail preservation and computational lightweighting. The MCA (Multi-Channel Attention) module enhances the discriminative semantic features between vegetation and bare soil via multi-scale pooling and an attention mechanism. The feature fusion module effectively bridges the semantic level gap between the encoder and decoder, enabling efficient aggregation of global semantics and local details.

Ultimately, CWSSNet achieves IoU values of 79.45\% for vegetation and 85.63\% for bare soil in classification, significantly outperforming other comparative methods. The results demonstrate that CWSSNet can effectively address the misclassification problems caused by complex morphologies and similar spectral characteristics. While improving the ability to distinguish highly similar land cover types, it also enhances the accuracy and detail coherence of semantic segmentation in complex scenarios.

\subsection{Ablation Experiments}
To evaluate the contribution of each module in CWSSNet, this study designed an ablation experiment scheme as shown in Table 2. Among the schemes, Experiment Group 0 replaces the WTBC module with a depthwise separable convolution block of the same receptive field size based on the baseline model. The classification results of each experimental group are presented in Table 3: the mIoU of Experiment Group 0 is 74.77\%, which is 0.22\% higher than that of the baseline. This indicates that the introduction of global contextual information can enhance the model’s ability to understand semantics in complex scenarios.  

Under the condition that the receptive field size (denoted as \( R \)) is the same, the difference in the number of parameters between the WTBC module and the standard depthwise convolution can be quantified through theoretical derivation. Let the number of input channels be \( C_{\text{in}} \), and the convolution kernel size of the standard depthwise convolution be \( k_{\text{std}} = R \); its number of parameters is expressed as:

\begin{equation}
P_{\text{std}} = k_{\text{std}}^2 \cdot C_{\text{in}} = R^2 \cdot C_{\text{in}}
\label{eq:p_std}
\end{equation}

% 文档正文中的表格代码
\begin{table}[htbp]
  \centering
  \setlength{\extrarowheight}{4pt}  % 增加行间距，避免内容拥挤（可选）
  % 列格式：8列居中对齐，每列两侧加竖线（|c| 表示带竖线的居中列）
  \begin{tabular}{cccccccc}
    \hline  % 顶部横线
    No.            & 0 & 1 & 2 & 3 & 4 & 5 & 6 \\
    \hline
    MCA            & $\checkmark$ & $\checkmark$ &  & $\checkmark$ & $\checkmark$ &  &  \\

    WTBC           & $\checkmark$ & $\checkmark$ & $\checkmark$ &  &  & $\checkmark$ &  \\

    feature fusion & $\checkmark$ &  & $\checkmark$ & $\checkmark$ &  &  &  \\
    \hline  % 底部横线
  \end{tabular}
  \caption{Settings of Ablation Experiments}
  \label{tab:method_feature}
\end{table}

\begin{table}[]
\caption{Results of Ablation Experiments}
\centering
  \setlength{\extrarowheight}{4pt}  % 增加行间距，避免内容拥挤（可选）
  % 列格式：8列居中对齐，每列两侧加竖线（|c| 表示带竖线的居中列）
\begin{tabular}{cccccccc}
\hline
No. & Built-up Area & Water & Vegetation & Bare Land & Road  & Other & mloU  \\
\hline
0   & 86.09         & 91.87 & 79.45      & 85.63     & 45.54 & 71.25 & 74.77 \\
1   & 85.18         & 90.81 & 76.56      & 84.58     & 44.63 & 71.16 & 74.20 \\
2   & 85.26         & 91.33 & 75.29      & 84.83     & 44.04 & 71.21 & 74.13 \\
3   & 84.95         & 89.86 & 74.73      & 83.02     & 44.38 & 71.05 & 73.63 \\
4   & 69.19         & 79.46 & 69.10      & 78.73     & 40.30 & 65.23 & 64.19 \\
5   & 68.16         & 78.26 & 69.17      & 76.08     & 40.28 & 63.12 & 63.58 \\
6   & 67.25         & 77.13 & 68.23      & 74.27     & 39.59 & 62.35 & 61.84 \\
\hline
\end{tabular}
\end{table}

The WTBC module adopts a wavelet multi-level decomposition strategy. Let the decomposition level be \( L \) and the size of the single-level convolution kernel be \( k \); its equivalent receptive field is \( R = 2L \cdot k \) (the receptive field doubles with each level of decomposition). Each level of wavelet decomposition generates 4 sub-bands (including 1 low-frequency sub-band and 3 high-frequency sub-bands); therefore, the total number of parameters for \( L \)-level decomposition is expressed as:

\begin{equation}
P_{\text{WTBC}} = L \cdot 4 \cdot k^2 \cdot C_{\text{in}} = L \cdot 4 \cdot \left(\frac{R}{2^L}\right)^2 \cdot C_{\text{in}} = \frac{L}{4^{L-1}} \cdot R^2 \cdot C_{\text{in}}
\label{eq:p_wtbc}
\end{equation}

The ratio of their parameter counts is expressed as:

\begin{equation}
\frac{P_{\text{WTBC}}}{P_{\text{std}}} = \frac{L}{4^{L-1}}
\label{eq:p_ratio}
\end{equation}

This ratio decreases exponentially as the number of decomposition levels \( L \) increases. This indicates that the WTBC module achieves a significant reduction in the number of parameters while maintaining an equivalent receptive field, and the higher the decomposition level, the more pronounced the advantage in parameter efficiency. A specific comparison is shown in Figure 7: when using 2-level db2 wavelets, the number of parameters of the WTBC module is already significantly lower than that of the standard convolution.  

To further enhance the spatial continuity of the features output by the MCA module and adapt to the characteristics of remote sensing images—such as diverse land cover scales and complex boundaries—the WTBC module adopts two convolution kernel sizes (3×3 and 5×5) for extracting local details and medium-range features, respectively. To verify the effectiveness of this design, comparative experiments with different convolution kernel configurations were conducted, and the results are presented in Table 4.

The experiments show that the 3×3 convolution performs better for small-scale land covers (e.g., construction land), while the 5×5 convolution is more suitable for modeling medium-scale targets (e.g., vegetation). The multi-scale convolution kernel design combining the two achieves optimal performance across multiple evaluation metrics, with the mIoU increased to 74.77\%. This verifies the effectiveness of the multi-scale feature extraction strategy.

\begin{figure} % picture
\caption{Comparison between WTBC and Standard Convolution}
    \centering
    \includegraphics[scale=0.7]{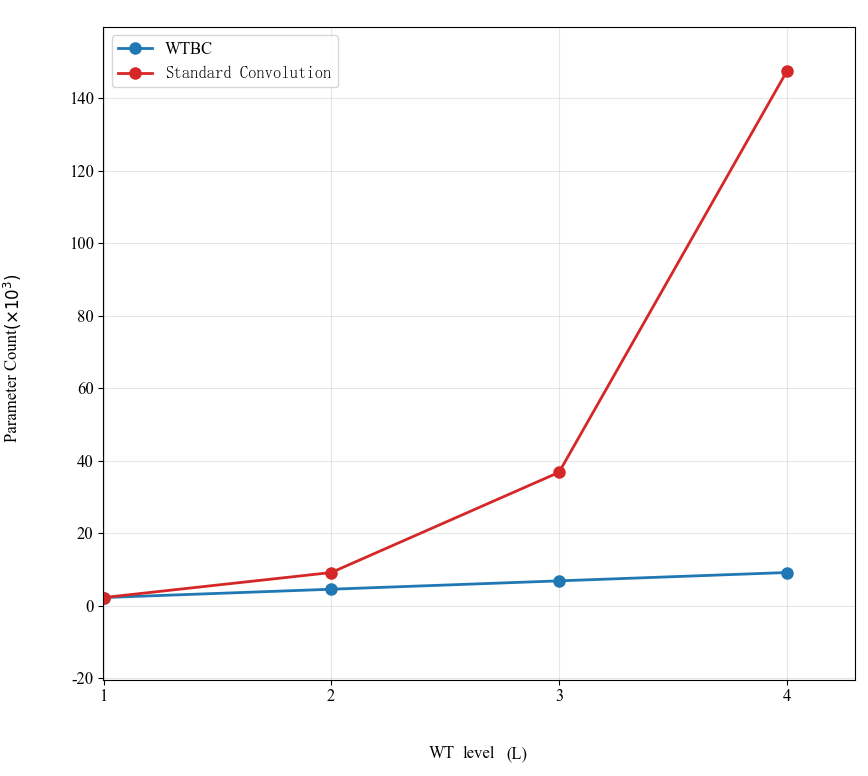}
\end{figure}

\begin{table}[]
\caption{Results of Ablation Experiments with Different Convolution Kernel Combinations}
\centering
  \setlength{\extrarowheight}{4pt}  % 增加行间距，避免内容拥挤（可选）
  % 列格式：8列居中对齐，每列两侧加竖线（|c| 表示带竖线的居中列）
\begin{tabular}{cccccccc}
\hline
convolution kernel size                                  & Built-up Area & Water & Vegetation & Bare Land & Road  & Other & mloU  \\
\hline
3×3                                                      & 85.95         & 90.84 & 76.55      & 84.46     & 44.25 & 71.24 & 74.32 \\
5×5                                                      & 85.88         & 90.81 & 76.92      & 84.38     & 44.64 & 71.26 & 74.13 \\
\begin{tabular}[c]{@{}c@{}}3×3+5×5\end{tabular} & 86.09         & 91.87 & 79.45      & 85.63     & 45.54 & 71.25 & 74.77\\
\hline
\end{tabular}
\end{table}

\section{Conclusions}  
To address the issues of feature redundancy and limited receptive fields in hyperspectral image land cover classification, this study proposes a CWSSNet model that integrates wavelet transform and attention mechanisms. Through systematic experiments and analysis, the following conclusions are drawn:  

CWSSNet achieves an mIoU of 74.50\% in the Yugan County experimental area, representing an improvement of 0.92\% to 1.42\% compared with various comparative methods (including Decision Tree, SVM, RF, 2D-CNN, 3D-CNN, and HybridSN). It performs particularly well in hard-to-distinguish classes such as water bodies (91.87\%), vegetation (79.45\%), and bare soil (85.63\%). This proves the effectiveness of the collaborative optimization strategy combining multi-scale frequency-domain features and spatial attention mechanisms.  

The WTBC module realizes the exponential expansion of the receptive field (the expansion degree increases with the number of wavelet decomposition levels \( L \)) with fewer parameters through wavelet multi-scale decomposition and binary convolution operations. When \( L = 2 \), the number of parameters of this module is reduced by 37.2\% compared with standard convolution. Meanwhile, the multi-scale convolution kernel combination (3×3 and 5×5) brings an mIoU improvement of 0.44\% to 0.64\% compared with the single convolution kernel setting, verifying its adaptability and superiority in handling multi-scale land cover features.

\bibliographystyle{unsrt}  


\end{document}